%% file: main.tex
\documentclass[10pt,twocolumn,letterpaper]{article}

\usepackage[pagenumbers]{cvpr} 

\input{preamble}
\definecolor{cvprblue}{rgb}{0.21,0.49,0.74}
\usepackage[pagebackref,breaklinks,colorlinks,allcolors=cvprblue]{hyperref}
\usepackage{amssymb}
\usepackage{amsmath}
\usepackage{graphicx}
\usepackage{stfloats}
\usepackage{float}
\usepackage{booktabs}
\usepackage{wrapfig}
\usepackage{multirow}
\usepackage{subcaption}
\usepackage{pifont}
\usepackage{enumitem}
\usepackage[table]{xcolor}
\newcommand{\TopOne}[1]{\cellcolor{blue!35}{#1}}
\newcommand{\TopTwo}[1]{\cellcolor{blue!20}{#1}}
\newcommand{\TopThree}[1]{\cellcolor{blue!10}{#1}}
\newcommand{\GOne}[1]{\cellcolor{green!35}{#1}}  
\newcommand{\GTwo}[1]{\cellcolor{green!20}{#1}}  
\newcommand{\GThree}[1]{\cellcolor{green!10}{#1}} 
\usepackage{tabularx,array,booktabs,multirow}
\newcolumntype{Y}{>{\centering\arraybackslash}X} 
\usepackage{array}      
\usepackage[linesnumbered,ruled,vlined]{algorithm2e}
\SetKwInput{KwRequire}{Require}
\SetKwInput{KwEnsure}{Ensure}
\SetKw{Return}{return}

\def\paperID{17641} 
\def\confName{CVPR}
\def\confYear{2026}

\title{VISTAv2: World Imagination for Indoor Vision‑and-Language Navigation}

\author{
Yanjia Huang \quad
Xianshun Jiang \quad
Xiangbo Gao \quad
Mingyang Wu \quad
Zhengzhong Tu\thanks{Corresponding author: \texttt{tzz@tamu.edu}}\\
Texas A\&M University, TACO Group\\
\url{https://taco-group.github.io/}
}

\begin{document}
\maketitle
\input{sec/0_abstract}    
\input{sec/1_intro}

\input{sec/2_related_works}

\input{sec/3_method}

\input{sec/4_experiments}
\input{sec/5_conclusion}
\input{sec/6_limitation}

\newpage
{
    \small
    \bibliographystyle{unsrtnat}
    \bibliography{main}
}


\end{document}

%% file: sec/0_abstract.tex
\begin{abstract}

Vision-and-Language Navigation (VLN) requires agents to follow language instructions while acting in continuous real world spaces. Prior image imagination based VLN work shows benefits for discrete panoramas but lacks online, action-conditioned predictions and does not produce explicit planning values; moreover, many methods replace the planner with long-horizon objectives that are brittle and slow. To bridge this gap, we propose \textbf{VISTAv2}, a generative world model that rolls out egocentric future views conditioned on past observations, candidate action sequences, and instructions, and projects them into an online value map for planning. Unlike prior approaches, VISTAv2 does not replace the planner. The online value map is fused at score level with the base objective, providing reachability and risk-aware guidance. Concretely, we employ a Conditional Diffusion Transformer video predictor aware of action to synthesize short-horizon futures, align them with the natural language instruction via a vision-language scorer, and fuse multiple rollouts in a differentiable Imagination-to-Value head to output an imagined egocentric value map. For efficiency, rollouts occur in VAE latent space with a distilled sampler and sparse decoding, enabling inference on a single consumer GPU. Evaluated on MP3D and RoboTHOR, VISTAv2 improves over strong baselines, and ablations show that action-conditioned imagination, instruction-guided value fusion, and the online value-map planner are all critical—suggesting that VISTAv2 offers a practical and interpretable route to robust VLN. 
\end{abstract}

%% file: sec/1_intro.tex
\section{Introduction}
\label{sec:intro}
    \textit{“The map is not the territory.”}
\hfill - Alfred Korzybski

Enabling embodied agents to understand natural instructions and locate them both quickly and robustly in real-world environments without prior maps still remains a core challenge in Vision-and-Language Navigation tasks \cite{xin2018reinforced, jinyu2022reinforced, r2r-dataset, wu2024voronavvoronoibasedzeroshotobject, wang2019reinforcedcrossmodalmatchingselfsupervised}. There are two different methods for approaching this, each with strengths and limitations. One relies on Large Vision-Language Models (VLMs) to match "currently observed objects" with "verbalized goals" through scoring the similarity, then employs frontier exploration or occupancy maps for "semantically driven mapping and search" (e.g., VLFM \cite{yokoyama2023vlfmvisionlanguagefrontiermaps}). The other leverages world models to perform long-range rollouts in action space, evaluating path quality by "imagining future scenes" (e.g., Navigation World Model and VISTA \cite{bar2025navigationworldmodels, huang2025vistagenerativevisualimagination}). The former exhibits strong semantic generalization and simple deployment but lacks explicit evaluation of reachability and geometric information, often leading to looks right but turns out wrong detours. The latter can explicitly assess evidence of reachability but suffers from long-rollout fragility which makes it challenging to achieve stable gains.

In this paper, we argue that VLN benefits from \emph{short-horizon, action-conditioned imagination that lives in map space}. We observe the followings. Pure language–vision matching tells what looks relevant but not whether it is reachable from the current pose. It cannot foresee occlusions or near-term collisions, and often over-scores visually salient but blocked targets (e.g., behind walls, glass, or outside the current room), leading to detours and backtracking \cite{dhruv2022lmnav, chenguang2022visual, arun2019talk2nav, jiazhao2024navid, xin2018look, shizhe2022think}. Optimizing a full world model objective over long-horizons amplifies small pose errors and sampling noise; appearance drift across dozens of denoising steps causes the planner to chase artifacts rather than geometry. The compute budget required for long rollouts also reduces the number of candidates we can evaluate per step \cite{matthew2023goat, khanh2018visionbased, jialu2023panogen, meng2025streamvln, zhiyuan2024cogga}. 

To address these challenges, we present \textbf{VISTAv2}, a language-conditioned, action-aware generative world model. VISTAv2 \underline{(i)} rolls out short-horizon egocentric futures, \underline{(ii)} turns them into an imagined value map by combining instruction–vision alignment with traversability and obstacle cues, and \underline{(iii)} re-ranks planner candidates via score-level fusion with the base objective. Overall, VISTAv2 is a test-time plug-in to standard frontier-based exploration planners, runs online through latent-space diffusion with distillation, and shows improvements on VLN datasets. 


Our contributions are summarized as follows:
\begin{itemize}
  \item A world model for VLN that performs \emph{short-horizon, action-conditioned} rollouts and expresses their guidance as an \emph{egocentric value map} in map space.
  \item An \emph{Imagination-to-Value} head that converts predicted futures into a value map and guides planning via score fusion, without replacing the planner.
  \item         Compared to VISTA, VISTAv2 improves Val-Unseen SR and SPL by +3.6 and +5.4 with shorter TL (13.26$\rightarrow$10.73, \(-19\%\)), and improves Test-Unseen SPL by +2.3 with shorter TL (14.20$\rightarrow$12.44, \(-12\%\)).
\end{itemize}

\begin{figure*}[t]
    \centering
    \includegraphics[width=1\linewidth]{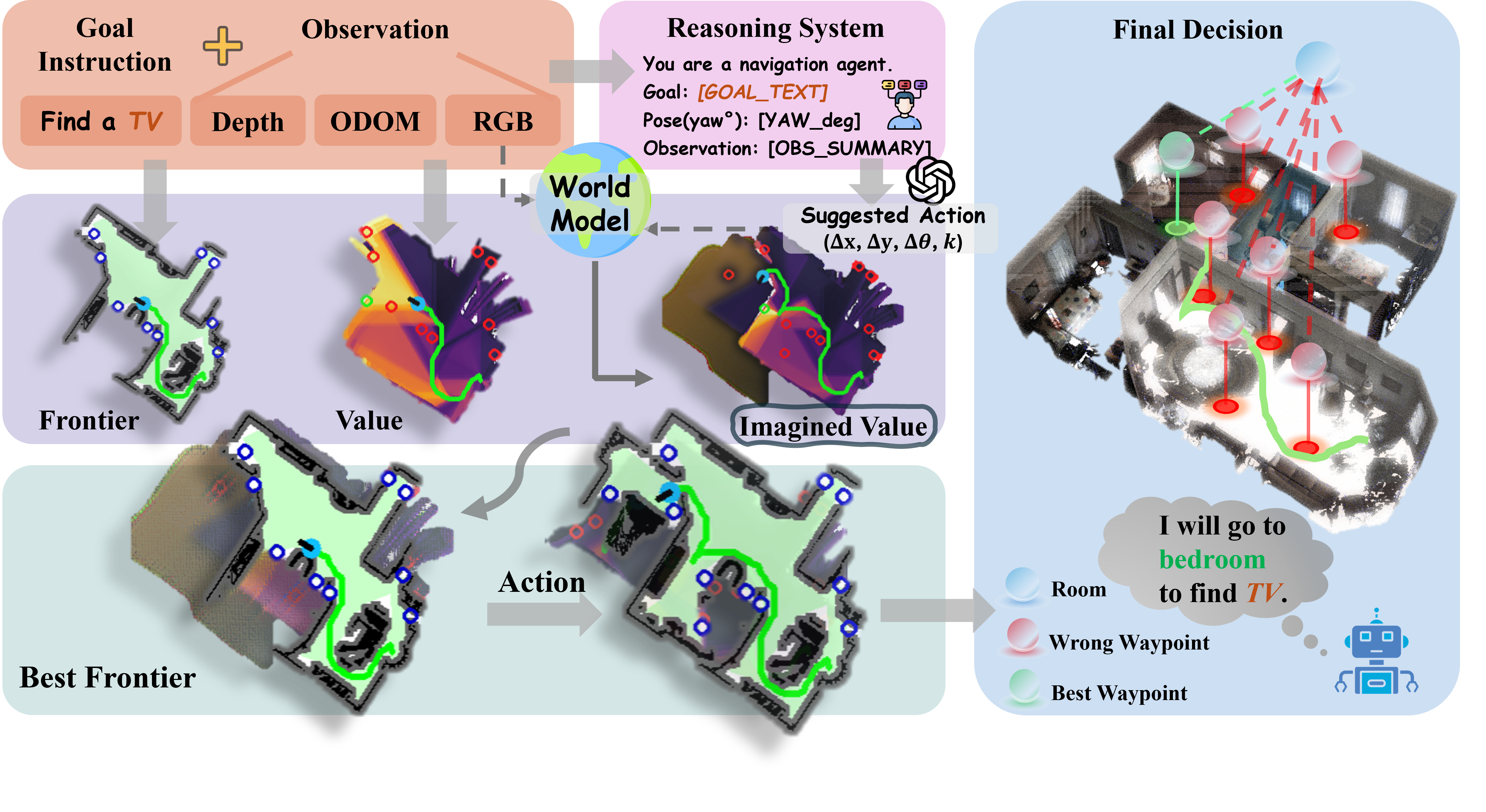}
\caption{\textbf{VISTAv2 pipeline overview (\S\ref{sec3.1}).} 
From a language instruction and observations (RGB, depth, odometry), the agent: 
(1) builds a local map and proposes frontier-based candidate trajectories; 
(2) forms a language prior over the map (Value); 
(3) uses the world model to imagine short-horizon futures and converts them into an egocentric imagined value map; 
(4) fuses imagined value and the prior with the planner’s native score to rank candidates (Eq.~\eqref{eq:fusion-final}) and executes the first control in a receding-horizon loop.}
    \label{Figure 1}
\end{figure*}

%% file: sec/2_related_works.tex
\section{Related Work}
\label{sec:Related Work}
\paragraph{Vision-and-Language Navigation (VLN).}
VLN links natural language instructions with embodied perception and control \cite{ganlong2024overnav, vlnce, yibo2023grounded, chiawen2022structureencoding, sonia2021languagealigned, tachung2019just, ronghang2019are, dillon2024adavln, yicong2020subinstruction, VLT}. Early work posed the problem on panoramic navigation graphs with R2R and RxR \cite{r2r-dataset, rxr-dataset}, later enabled at scale by the Habitat simulator \cite{habitat-simulator} for photo-realistic, high speed training and evaluation. And beyond navigation graphs, VLN-CE lifts agents into continuous 3D control with egocentric sensing and realistic collisions, removing graph constraints and exposing geometric challenges central to real robots \cite{krantz2020navgraphvisionandlanguagenavigationcontinuous}. Recent trends leverage foundation or VLM models to strengthen language grounding and zero-shot generalization \cite{huang2025vistagenerativevisualimagination, lin2025navcot, jialu2023panogen, Wang_2025, meng2025streamvln, zhan2024mcgptempoweringvisionandlanguagenavigation, kurita2020generativelanguagegroundedpolicyvisionandlanguage}, complementary work studies injecting visual imagination as an added modality to provide landmark cues, yielding measurable SR gains on VLN agents. However, current VLN systems still struggle to reason about reachability and collision risk, VLM only scoring often favors visually salient yet unreachable directions, and long-horizon rollouts can be brittle and slow, compounding errors in continuous settings.



\paragraph{Video Generation as World Models.} 
Video generative models can act as controllable world models that roll out egocentric futures conditioned on past observations and actions. Recent systems show that such models can steer planning or augment downstream reasoning, either by playing out imagined trajectories or by ranking candidate plans using synthesized evidence \cite{bar2024lumiere, girdhar2023emu, ho2022imagen, tulyakov2018mocogan, yu2023magvit}. Notable examples include large, action-conditioned controllable world models like Genie \cite{bruce2024geniegenerativeinteractiveenvironments}, which could generate interactive environments from both image or text prompts and allow agent control via input actions.  Navigation World Models instantiate this idea for visual navigation with a controllable video generator that predicts future observations from past frames and navigation actions \cite{bar2025navigationworldmodels}. The model is trained on diverse egocentric videos and is used after training to simulate trajectories and verify whether a candidate plan reaches the goal. NWM is built on a Conditional Diffusion Transformer (CDiT), whose complexity is linear in the number of context frames, enabling scaling to large models while maintaining strong prediction quality. At deployment, NWM uses the generator in MPC and re-ranking for both insides and outsides of rollouts. This design reduces exposure to long-horizon drift without rewriting the planner.

\paragraph{Language Guided Mapping and Frontier Exploration.}
This line of work steers exploration by aligning egocentric observations with instruction text and projecting the scores onto maps or frontiers. A representative example is VLFM, which builds a frontier map and ranks candidate frontiers by vision–language similarity to the goal text, enabling zero-shot semantic navigation without task-specific training~\cite{yokoyama2023vlfmvisionlanguagefrontiermaps}. Follow up works extend the idea with panoramic cues and imagination from discrete viewpoints \cite{jialu2023panogen}, further improving frontier selection under language guidance~\cite{Wang_2025}. These methods provide strong semantic priors and simple deployment, yet they reason about reachability and collision risk only indirectly, high-scoring regions can be occluded or geometrically infeasible, and the scores are typically static \cite{laina2025findanythingopenvocabularyobjectcentricmapping, uno2025lgrllmguidedrankingfrontiers, jiang2025dualmaponlineopenvocabularysemantic, Yu_2023, chenguang2022visual}. Our work complements this trend by injecting an \emph{action conditioned, short-horizon} prior: we convert imagined futures into an egocentric value map and fuse it with the planner’s native objective at score level, preserving the underlying map-based search while adding instruction consistent, geometry-aware guidance.

\begin{figure*}[t]
    \centering
    \includegraphics[width=1\linewidth]{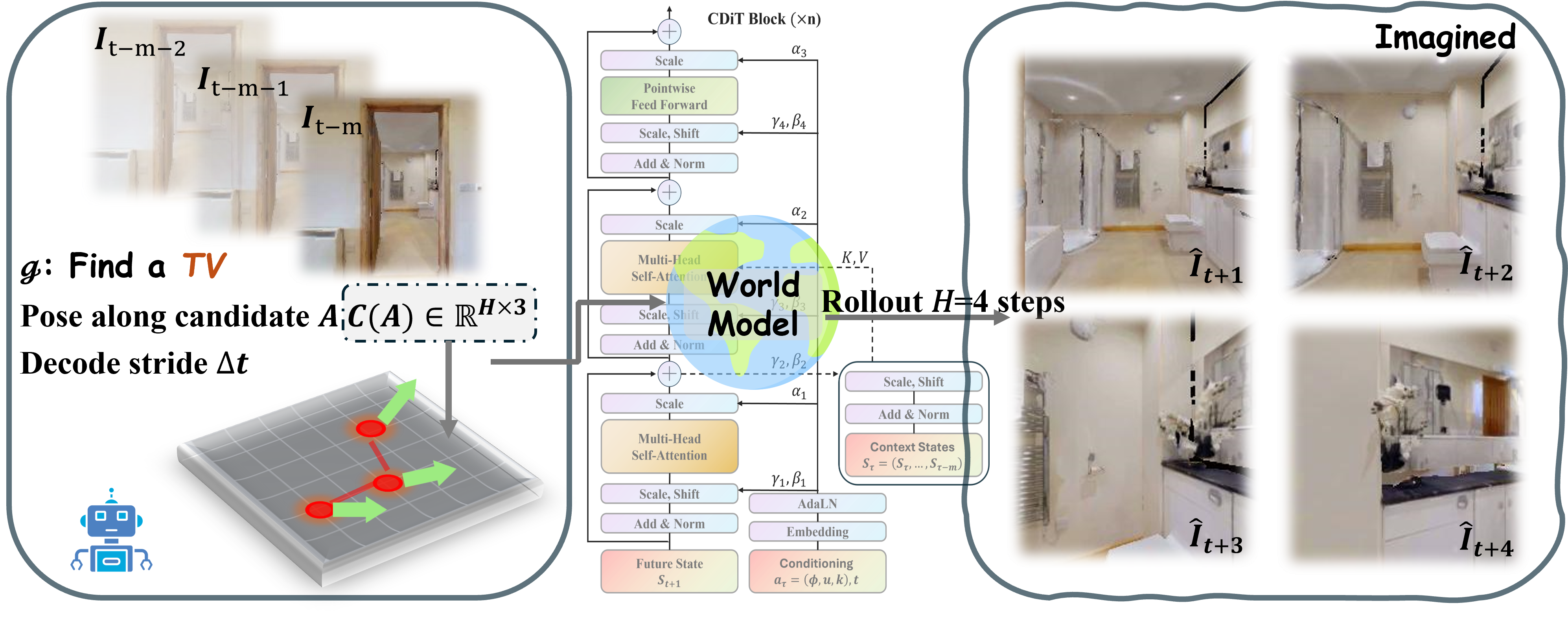}
\caption{\textbf{World Model (\S\ref{sec3.2}).} 
Given the recent egocentric frames $I_{t-m+1:t}$, the instruction $g$, and a candidate trajectory $A$, we integrate poses to obtain $C(A)\!\in\!SE(2)^H$ and feed $(x_t,g,C(A))$ to the action-conditioned video diffusion model $\mathcal{W}_\theta$ (CDiT in VAE latent space). 
$\mathcal{W}_\theta$ produces a short-horizon egocentric rollout $\{\hat{I}_{t+\tau}\}_{\tau=1}^{H}$; only a stride-$\Delta t$ subset is decoded for downstream I2V scoring (\S\ref{sec3.3}).}

    \label{Figure 2}
\end{figure*}

%% file: sec/3_method.tex
\section{Method}

We target \emph{test-time enhancement} for Vision-and-Language Navigation in standard VLN environments: at inference, an agent augments its default policy with a language-conditioned, action aware generative world model that imagines future egocentric observations and converts them into an online value map for planning (Fig.~\ref{Figure 1}). Our approach, VISTAv2, couples two components tightly: \ding{182} a video diffusion \textbf{world model} that rolls out short-horizon egocentric futures conditioned on the instruction and candidate actions; \ding{183} an \textbf{Imagination-to-Value (I2V)} head converts the imagined futures into an egocentric value map. At test-time, we do not replace the planner: the value map is fused at score level with the planner’s native objective to rank candidate actions, keeping the search procedure intact.

Sec.\ref{sec3.1} overviews the VISTAv2 pipeline. Sec.\ref{sec3.2} formalizes the language-conditioned, action-aware world model. Sec.\ref{sec3.3} introduces the I2V head, which converts multi rollouts, instruction aligned signals into an egocentric value map and fuses it with the base planner’s score for action selection. Finally, Sec.\ref{sec3.4} details the architecture and training setup used in our experiments.

\subsection{Pipeline Overview}
\label{sec3.1}
Fig.~\ref{Figure 1} illustrates our pipeline: given a language instruction $g$ and egocentric observations at time $t$ (RGB $I_t$, depth $D_t$, odometry $P_t$), VISTAv2 augments a standard frontier-based navigation stack with action-conditioned imagination and score-level fusion. Four components participate: a \emph{World Model} $\mathcal{W}$, a \emph{Language Reasoner} $\mathcal{R}$ (text–vision scorer), an \emph{Imagination-to-Value} head $\mathcal{H}$, and a base \emph{Planner} $\mathcal{P}$. 

Given egocentric $I_t$, $P_t$, and an instruction goal $g$, we launch an $n$-step candidate–expansion loop instead of committing to an action directly. For each trajectory in the current candidate set such as frontiers with short-horizon paths, the world model rolls out $H$-step egocentric futures (we take $H$=4 here), yielding imagined observations. Conditioned on the instruction, a language–vision scorer (BLIP \cite{li2022blip}) evaluates these imagined observations and (i) writes instructive evidence into a small buffer, and (ii) via an Imagination-to-Value (I2V) head converts them into an egocentric \textbf{imagined} value map. We fuse the imagined value with the planner’s native score to rank candidates and keep the top-$K$ to form the next layer. After the search, the planner executes the first control of the best candidate; mapping updates and the loop repeats at the next step. This \emph{imagine $\rightarrow$ score $\rightarrow$ fuse $\rightarrow$ act} procedure augments a frozen navigation stack with world model priors and motion forecasts, improving VLN at test-time.

\subsection{World Model Formulation}
\label{sec3.2}
We treat the world model as an egocentric video generation simulator that rolls out a sequence of actions from previous observations. Given the last $m$ RGB frames $I_{t-m+1:t}$, poses $P_{1:t}$, an instruction $g$, and a candidate control sequence $A_{t:t+H-1}$, the model predicts short-horizon egocentric futures that will later be converted into an imagined value map (Sec.~\ref{sec3.3}).

\paragraph{Action Space.}
We use a continuous, egocentric control space
\[
\mathcal{A}\subset\mathbb{R}^{4},\quad
a_\tau=(\Delta x_\tau,\,\Delta y_\tau,\,\Delta\theta_\tau,\,\kappa_\tau),
\]
where $(\Delta x,\Delta y)$ are planar displacements in the local frame, $\Delta\theta$ is the yaw change, and $\kappa$ scales step duration. Feasibility is enforced by platform limits
$\|(\Delta x,\Delta y)\|\le v_{\max}\kappa\Delta t$ and $|\Delta\theta|\le \omega_{\max}\kappa\Delta t$,
and a \texttt{STOP} primitive is available when appropriate.

\paragraph{Action Representation.}
A candidate $A_{t:t+H-1}$ from the base planner is resampled to a fixed horizon $H$.
Each step is embedded as
\[
\tilde a_\tau=\Big[\tfrac{\Delta x_\tau}{\sigma_x},\,\tfrac{\Delta y_\tau}{\sigma_y},\,\sin\Delta\theta_\tau,\,\cos\Delta\theta_\tau,\,\tfrac{\kappa_\tau}{\sigma_\kappa}\Big],
\]
concatenated with a sinusoidal time code and a learned mode bit (turn vs. go-straight). The sequence
$\tilde A_{t:t+H-1}$ is provided as per-timestep conditioning tokens.


\paragraph{Action-Driven Video Generation.}
Let $x_t\in\mathbb{R}^{H\times W\times 3}$ be the current egocentric frame and
$C(A)=\big(c_1,\dots,c_H\big)\in SE(3)^H$ the pose sequence induced by a candidate trajectory $A$ starting from pose $P_t$ (for ground robots $C(A)\subset SE(2)$).
Our instruction- and pose-conditioned video diffusion model
\begin{equation}
\label{eq:world-model}
\begin{aligned}
\mathcal{W}_\theta:\;(x_t,g,C(A)) \mapsto\; \mathbf{X}(A)
  &= \{\hat{x}_{t+\tau}\}_{\tau=1}^{H},\\[-2pt]
\hat{x}_{t+\tau} &\in \mathbb{R}^{H\times W\times 3}.
\end{aligned}
\end{equation}
maps the triplet $(x_t,g,C(A))$ to an \emph{egocentric rollout} that follows the intended motion while emphasizing instruction-relevant visual evidence.
In practice we operate in latent space, i.e., with $z_t=E(x_t)$ and $\hat{z}_{t+\tau}$ rolled out by a CDiT; only a sparse subset is decoded $\hat{x}_{t+\tau}=D(\hat{z}_{t+\tau})$ for downstream scoring.

\paragraph{Language Reasoner and Prior Map.}
We use a frozen vision–language encoder $\mathcal{R}$ (BLIP) to obtain a text embedding $\phi_g$ and per-pixel/patch image features $\phi(\cdot)$. 
Given the current observation set and the local map $\mathcal{M}_t$, we compute a \emph{language prior} $V_t^{\text{prior}}$ by projecting alignment scores to the egocentric grid:
\[
\begin{aligned}
s(u,v) &\!=\! \mathrm{softmax}_\tau\!\big(\cos\!\langle \phi_g,\,\phi(I_t)[u,v]\rangle\big),\\
V_t^{\text{prior}} &\!=\! \Pi\!\big(s(\cdot)\rightarrow\mathcal{M}_t\big).
\end{aligned}
\]

where $\Pi(\cdot)$ denotes depth-aware splatting onto the map and $\tau$ is a temperature. 
$V_t^{\text{prior}}$ captures instruction relevance but is action-agnostic; it is fused at score level in Eq.~\eqref{eq:fusion-final}.

\paragraph{Score-level fusion.}
For a candidate $A_{t:t+H-1}$, we fuse imagined value and the planner’s native score:

\begin{equation}
\label{eq:fusion-final}
\begin{aligned}
S_{\text{fused}}\big(A_{t:t+H-1}\big)
&= S_{\text{base}}\big(A_{t:t+H-1}\big)\\
&\quad+ \lambda_1 \sum_{\tau=1}^{H}\gamma^{\tau}\,
   V_t^{\text{img}}\!\big(x_\tau(A),y_\tau(A)\big)\\
&\quad+ \lambda_2 \sum_{\tau=1}^{H}\gamma^{\tau}\,
   V_t^{\text{prior}}\!\big(x_\tau(A),y_\tau(A)\big).
\end{aligned}
\end{equation}

where $(x_\tau(A),y_\tau(A))$ are egocentric samples along $A$, $\gamma\!\in\!(0,1]$ is a temporal discount, and $\lambda_1,\lambda_2\!\ge\!0$ are fusion weights. The highest-scoring candidate is chosen and only its first control is executed (receding horizon).

\SetKwFunction{IntegratePoses}{IntegratePoses}
\SetKwFunction{EmbedActions}{EmbedActions}
\SetKwFunction{ProjectAndFuse}{ProjectAndFuse}
\SetKwFunction{Append}{append}
\SetKwFunction{Wstep}{W_{\theta}.\,step}

\begin{algorithm}[t]
\caption{$\mathcal{W}_\theta$.\textsc{Rollout} — Action-Driven Video Generation}
\label{alg:worldmodel}
\KwIn{$I_{t-m+1:t}$, $P_t$, instruction $g$, candidate $A_{t:t+H-1}$, horizon $H$, decode stride $\Delta t$, denoise steps $K_d$}
\KwOut{$C(A)=(c_1,\dots,c_H)$, latent rollout $\hat{z}_{t+1:t+H}$, decoded frames $\mathbf{X}(A)=\{\hat{x}_{t+\tau}\}_{\tau\in\mathcal{S}}$}
$C(A)\leftarrow \textsc{IntegratePoses}(P_t, A_{t:t+H-1})$; \quad $\mathcal{S}\leftarrow\{\,\tau:\tau=1,1{+}\Delta t,\dots,H\,\}$\;
$z_{t-m+1:t}\leftarrow E(I_{t-m+1:t})$; \quad $\hat{z}_{t+1:t+H}\leftarrow \emptyset$; \quad $\mathbf{X}(A)\leftarrow \emptyset$\;
\For{$\tau=1$ \KwTo $H$}{
  $\text{cond}\leftarrow \big(z_{t-m+1:t},\,\hat{z}_{t+1:t+\tau-1},\,\Phi(g),\,\tilde{A}_{t:t+\tau-1},\,\tau\big)$\;
  $\hat{z}_{t+\tau}\leftarrow \textsc{CDiT\_Denoise}(\text{cond}; K_d)$\;
  \If{$\tau\in\mathcal{S}$}{
    $\hat{x}_{t+\tau}\leftarrow D(\hat{z}_{t+\tau})$; \quad $\mathbf{X}(A)\leftarrow \mathbf{X}(A)\cup\{\hat{x}_{t+\tau}\}$\;
  }
}
\Return $C(A)$, $\hat{z}_{t+1:t+H}$, $\mathbf{X}(A)$\;
\end{algorithm}

\subsection{Imagination to Value (I2V)}
\label{sec3.3}
\paragraph{Input.} For a candidate $A_{t:t+H-1}$, the world model (Sec.~\ref{sec3.1}) provides a sparse set of imagined egocentric frames $\{\hat{I}_{t+\tau}\}$, their poses $\{c_\tau\}$ and depth (predicted or monocular).

\paragraph{Frame scoring.} Each $\hat{I}_{t+\tau}$ is converted to a single confidence map by linearly combining three dense cues: (i) instruction alignment (text–vision similarity), (ii) traversability/free-space, and (iii) obstacle/uncertainty penalty. The weights are learned, nonnegative, and the output is temperature-scaled and normalized to $[0,1]$.

\paragraph{Egocentric projection.} Using camera intrinsics and pose $c_\tau$, the confidence map is softly projected onto a fixed egocentric grid around the agent (depth-aware splatting with bilinear accumulation). We apply a light morphological smoothing to account for the robot footprint and mask out regions outside the field of view.

\paragraph{Use.} $V_t^{\text{img}}$ is sampled along each candidate path and fused at score level with the planner’s native score (Eq.~\eqref{eq:fusion-final}). For efficiency, we decode every $\Delta t$ frames and use a compact grid (e.g., $80{\times}80$ within a $12$\,m window).

\paragraph{Multi-rollout aggregation.}
Given imagined frames at times $\tau\!\in\!\mathcal{S}$, we accumulate into the value map with discount $\gamma$:
\[
V_t^{\text{img}}(u,v)= \mathrm{LSE}_\beta\big\{\gamma^{\tau}\,\Pi(\tilde{v}_{t+\tau})(u,v)\big\}_{\tau\in\mathcal{S}},
\]
where $\mathrm{LSE}_\beta(\cdot)$ is a temperature-controlled log-sum-exp that smoothly approximates max-pooling and $\Pi$ is the egocentric projection. 
Using LSE avoids brittleness of hard maxima while preserving peakiness at promising regions.

\paragraph{Uncertainty gating.}
We estimate the rollout uncertainty $\sigma_A$ (from ensemble variance or diffusion variance) and disable imagination when it is high.
We use a hard gate with a fixed threshold $\theta\!=\!0.6$ across all experiments, see Figure \ref{Figure 5}.

\begin{algorithm}[t]
\caption{\textsc{OneStepPlan}: Imagine–Score–Fuse–Act (single step)}
\label{alg:vistav2}
\KwRequire{current $I_{t-m+1:t},D_{1:t},P_t$, map $\mathcal{M}_t$, instruction $g$; world model $\mathcal{W}$; reasoner $\mathcal{R}$; I2V head $\mathcal{H}$; base planner $\mathcal{P}$; horizon $H$, \#candidates $K$, discount $\gamma$, weights $\lambda_1,\lambda_2$, decode stride $\Delta t$}
\KwEnsure{next action $a_t$}
$\mathcal{B} \leftarrow \mathcal{P}.\textsc{Candidates}(\mathcal{M}_t,K)$ \tcp*[r]{frontiers $\rightarrow$ paths $A$}
$V_t^{\text{prior}} \leftarrow \mathcal{R}.\textsc{Prior}(g,\mathcal{M}_t,P_t)$
\For{\textbf{each} $A \in \mathcal{B}$}{
  $C(A) \leftarrow \textsc{IntegratePoses}(P_t,A)$ \tcp*[r]{poses along $A$}
  $\hat{\mathbf X}(A) \leftarrow \mathcal{W}.\textsc{Rollout}(I_{t-m+1:t},g,C(A);\Delta t)$
  $V_t^{\text{img}} \leftarrow \mathcal{H}.\textsc{I2V}(\hat{\mathbf X}(A),C(A))$
  $s_{\text{base}} \leftarrow \mathcal{P}.\textsc{Score}(A)$
  $s_{\text{img}} \leftarrow \textsc{SamplePath}(V_t^{\text{img}},A,\gamma)$
  $s_{\text{prior}} \leftarrow \textsc{SamplePath}(V_t^{\text{prior}},A,\gamma)$
  $S_{\text{fused}}(A) \leftarrow s_{\text{base}}+\lambda_1 s_{\text{img}}+\lambda_2 s_{\text{prior}}$ \tcp*[r]{Eq.~\eqref{eq:fusion-final}}
}
$A^\star \leftarrow \arg\max_{A \in \mathcal{B}} S_{\text{fused}}(A)$\;
\Return first control of $A^\star$\;
\end{algorithm}

\begin{table*}[t]
\centering
\caption{Performance on R2R. All methods use the 3\,m success radius and identical stop rules.}
\label{Table1}
\small
\renewcommand{\arraystretch}{1.05}
\setlength{\tabcolsep}{6pt}
\begin{tabularx}{\textwidth}{l *{8}{Y}}
\toprule
\multirow{2}{*}{\textbf{Models}} & \multicolumn{4}{c}{\textbf{Validation Unseen}} & \multicolumn{4}{c}{\textbf{Test Unseen}} \\
\cmidrule(lr){2-5}\cmidrule(lr){6-9}
& SR $\uparrow$ & SPL $\uparrow$ & TL $\downarrow$ & NE $\downarrow$
& SR $\uparrow$ & SPL $\uparrow$ & TL $\downarrow$ & NE $\downarrow$ \\
\midrule
EnvDrop~\cite{tan2019learning}         & 52.0  & 48.0  & \TopThree{10.70} & 5.22  & 51.0  & 47.0  & \TopTwo{11.66} & 5.23 \\
VLFM~\cite{yokoyama2023vlfmvisionlanguagefrontiermaps} & 52.5  & 30.4  & / & / & / & / & / & / \\
PREVALENT~\cite{hao2020towards}        & 58.0  & 53.0  & \TopTwo{10.19}  & 4.71  & 54.0  & 51.0  & \TopOne{10.51} & 5.30 \\
RecBERT~\cite{hong2020recurrent}       & 63.0  & 57.0  & 12.01           & 3.93  & 63.0  & 57.0  & 12.35          & 4.09 \\
HAMT~\cite{chen2021history}            & 66.2  & 61.5  & 11.46           & 3.62  & 65.0  & 60.0  & \TopThree{12.27} & 3.93 \\
HAMT\text-\!Imagine~\cite{perincherry2025visualimaginationsimprovevisionandlanguage}
                                       & 67.26 & 62.02 & 11.80           & 3.58  & 65.0  & 60.0  & 12.66          & 3.89 \\
DUET~\cite{chen2022think}              & 71.52 & 60.41 & 13.94           & 3.31  & 69.0  & 59.0  & 14.73          & \TopThree{3.65} \\
DUET-\!Imagine~\cite{perincherry2025visualimaginationsimprovevisionandlanguage}
                                       & 72.12 & 60.48 & 14.35           & \TopThree{3.19} & 71.0  & 60.0  & 15.35          & \TopTwo{3.52} \\
PanoGen~\cite{li2023panogentextconditionedpanoramicenvironment}
                                       & \TopThree{74.2} & 64.3  & 13.40 & \TopTwo{3.03} & \TopThree{71.7} & \TopThree{61.9} & 14.38 & \TopOne{3.31} \\
NavCoT~\cite{lin2025navcotboostingllmbasedvisionandlanguage}
                                       & 40.23 & 36.64 & \TopOne{9.95}   & 6.26  & / & / & / & / \\
OmniNav~\cite{xue2025omninavunifiedframeworkprospective}
                                       & 69.5  & \TopThree{66.1} & / & 3.74 & / & / & / & / \\
\midrule
VISTA~\cite{huang2025vistagenerativevisualimagination}
                                       & \TopTwo{77.8} & \TopTwo{68.3} & 13.26 & \TopOne{2.92} & \TopOne{74.9} & \TopTwo{66.7} & 14.20 & 3.77 \\
VISTAv2 (ours)                         & \TopOne{81.4} & \TopOne{73.7} & 10.73 & 3.76 & \TopTwo{73.1} & \TopOne{69.0} & 12.44 & 4.05 \\
\bottomrule
\end{tabularx}
\end{table*}

\subsection{World Model Details}
\label{sec3.4}

\paragraph{Architecture.}
Our world model uses a CDiT-L backbone , with the same encoder and decoder used throughout the Navigation World Model (NWM) \cite{bar2025navigationworldmodels}. Conditioning is injected by per-timestep action tokens and an instruction embedding, as described in Sec.~\ref{sec3.1}. All inputs are resized to $224\times224$ and processed in bf16 precision.

\paragraph{Training data.}
We curate \textbf{489k} egocentric RGB frames from HM3D and MP3D, organized as short trajectories with their pose sequences and actions. The train/val split follows our navigation splits; we use only indoor scenes and discard trajectories shorter than a fixed minimum length. Details and scene lists are provided in the Appendix.

\paragraph{Optimization and schedule.}
Unless otherwise noted, we train on 8$\times$~RTX~6000 Ada 48\,GB for $\sim$8 days, totaling $\approx\,$200k optimizer steps.
We use AdamW (lr $8\!\times\!10^{-5}$, weight decay $0.05$, $\beta{=}(0.9,0.999)$), cosine decay with 2k warmup, gradient clipping at 10.0, and an EMA with decay 0.9999.
Batch size is 8 with \emph{accumulation 2} (effective 16).
Mixed precision is bf16. We evaluate every 2k steps and save short rollouts for visual inspection.
Actions are $z$-score normalized over the training split.

%% file: sec/4_experiments.tex
\section{Experiments}

\subsection{Experiment Setup}
\label{Sec4.1}
\textbf{Benchmarks.}
We evaluate in Habitat on continuous VLN-CE episodes derived from the Room-to-Room corpus in Matterport3D (R2R). For R2R we follow the standard splits: train 10,819 episodes over 61 scenes, val-seen 778 over 53, val-unseen 1,839 over 11, and test 3,408 over 18 scenes \cite{r2r-dataset}. For HM3D, we adopt the public validation configuration with 2,000 episodes across 20 scenes and 6 goal categories to assess generalization in large-scale real-scan environments \cite{ramakrishnan2021habitatmatterport3ddatasethm3d}.

\textbf{Metrics.}
We report standard VLN metrics~\cite{ilharco2019generalevaluationinstructionconditioned}:
\ding{182} \emph{Success Rate (SR)} — fraction of episodes where the agent issues \textsc{Stop} within 3 meters of the goal;
\ding{183} \emph{Success weighted by Path Length (SPL)} — SR weighted by path efficiency relative to the shortest path;
\ding{184} \emph{Navigation Error (NE)} — shortest path distance (meters) from the final position to the goal;
\ding{185} \emph{Trajectory Length (TL)} — total distance traveled (meters).
Higher SR/SPL and lower NE/TL indicate better navigation. All inference is on a single GeForce RTX~4090; for VISTAv2 this includes world model rollouts, I2V scoring, and planner evaluation.

\begin{figure*}
    \centering
    \includegraphics[width=1\linewidth]{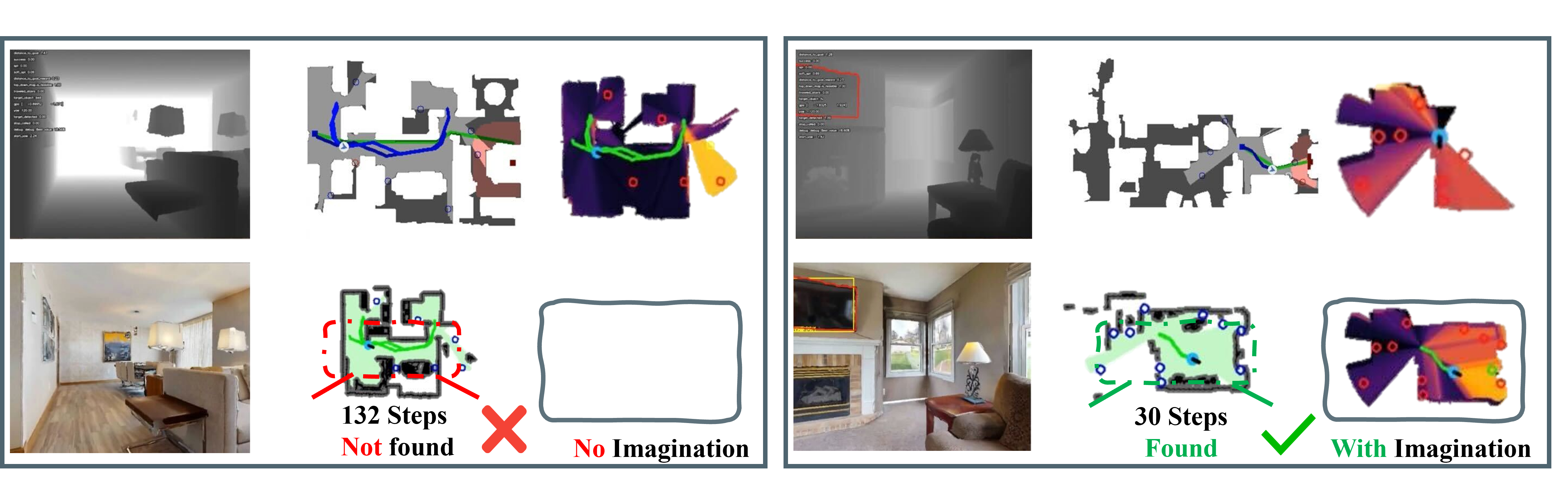}
    \caption{\textbf{Effect of visual imagination on goal discovery.} Each panel shows one episode (same start/goal). \textbf{Left (No Imagination):} the base planner explores many frontiers (red circles) guided only by occupancy/prior; the value over the explored region is diffuse (green mask), the agent wanders (\textbf{132 steps}) and fails to localize the TV. \textbf{Right (With Imagination):} our world model rolls out egocentric futures and the I2V head produces a fan-shaped image value map (orange/yellow), which fused with the prior sharpens the score and steers the agent through the doorway to the TV room, succeeding in 30 steps. Top: current depth/RGB and occupancy with path; Right column: value maps; Bottom: frontier set and fused score along the chosen path.}
    \label{Figure4}
\end{figure*}

\subsection{Experiments Results}
We evaluate VISTAv2 on both the R2R benchmark and RoboTHOR, and observe clear, consistent improvements over strong baselines across all splits in higher SR/SPL and with competitive NE/TL. 

\textbf{Baselines.}
We compare VISTAv2 with representative VLN/VLN-CE systems: 
EnvDrop, PREVALENT, RecBERT, HAMT, HAMT-Imagine, DUET, DUET-Imagine, VLFM, PanoGen(++), NavCoT, OmniNav, and the prior VISTA. When a method is defined on the panorama graph, we follow the standard R2R protocol and evaluate on the same splits; all methods share the same sensor suite and stopping rule for fair comparison.


\textbf{Main Results.}
Table~\ref{Table1} reports results on R2R (Val-Unseen/Test-Unseen). VISTAv2 yields higher SPL and shorter trajectories on both splits, with modest SR/NE trade-offs. Val-Unseen: 81.4 SR / 73.7 SPL (+3.6/+5.4 vs.\ VISTA), TL 13.26$\rightarrow$10.73 (–19\%), NE 3.76 vs.\ 2.92. Test-Unseen: 73.1 SR / 69.0 SPL (–1.8/+2.3), TL 14.20$\rightarrow$12.44, NE 4.05 vs.\ 3.77. Our fused policy tends to issue \textsc{Stop} once the imagined value concentrates near the goal and local confidence is high, which favors shorter, decisive paths yet can end just outside the 3\,m success radius.

\textbf{Imagination with value helps.}
Against semantic-only or frontier-style approaches (e.g., VLFM and PanoGen++ rows), VISTAv2 lifts SPL by \(\approx\!5-10\) points and SR by a similar margin, indicating that converting imagined futures into an \emph{egocentric value map} improves action selection beyond language–vision matching.

\textbf{Sequential imagination better than goal imagination.}
Compared to goal-only imagination baselines (e.g., HAMT-Imagine, DUET-Imagine), VISTAv2 yields large gains (Val: \(\uparrow\)SR by \(\approx\!9\)–\(\!14\) points; Test: \(\uparrow\)SR by \(\approx\!8\) points). We attribute this to our \emph{action-conditioned, short-horizon} rollouts and score-level fusion, which respect near-term reachability and geometric risk during planning. Overall, these trends support our claim: \emph{VLN agents perform better when imagination is expressed as a value in map space and used to re-rank candidates}, rather than relying on semantics alone or optimizing a long-horizon world-model objective.

{\setlength{\textfloatsep}{18pt}
\begin{table}[t]
\centering
\caption{Comparison on RoboTHOR.}
\label{tab:robothor_table}
\footnotesize                           
\renewcommand{\arraystretch}{1.05}       
\setlength{\tabcolsep}{3pt}              

\begin{tabularx}{0.95\columnwidth}{l *{2}{Y}}  
\toprule
\textbf{Model} & \textbf{SWPL$\uparrow$} & \textbf{SR$\uparrow$} \\
\midrule
CLIP-Ref.~\cite{samir2022cows}   & 2.4   & 2.7  \\
CLIP-Patch~\cite{samir2022cows}  & 10.6  & 20.3 \\
CLIP-Grad.~\cite{samir2022cows}  & 9.7   & 15.2 \\
MDETR~\cite{samir2022cows}       & 8.4   & 9.3  \\
OWL~\cite{samir2022cows}         & 17.2  & 27.5 \\
ESC~\cite{zhou2023escexplorationsoftcommonsense} & {22.2} & {38.1} \\
VLTNet~\cite{VLT}                & 17.1  & 33.2 \\
\midrule
VISTA~\cite{huang2025vistagenerativevisualimagination} & \TopTwo{28.8} & \TopTwo{43.1} \\
VISTAv2 (ours)                                        & \TopOne{34.7} & \TopOne{61.3} \\
\bottomrule
\end{tabularx}
\vspace{-2mm} 
\end{table}

\textbf{RoboTHOR Results.}
Table~\ref{tab:robothor_table} shows that VISTAv2 also brings sizable gains in the interactive RoboTHOR setting: it achieves 61.3 SR and 34.7 SWPL (Success-Weighted Path Length), outperforming VISTA by +18.2 SR and +5.9 SWPL, and surpassing CLIP-based, OWL, ESC, and VLTNet baselines by a large margin. The concurrent SR and SWPL gains suggest that map-space value priors from short-horizon rollouts both reduce outright failures and improve path efficiency, consistent with our R2R results.

We further tried to use the NWM to directly generate and score long-horizon future frames, but this naive pixel-space planner performed significantly worse due to severe rollout drift and artifacts.

\subsection{Ablations: From semantics to imagination to imagined value}
\label{sec:ablate_line1}

\textbf{Setup.} We contrast three stages: 
(1) \emph{VLFM} — frontier/map exploration scored only by vision–language similarity (no imagination);
(2) \emph{VISTA} — imagination for reasoning but \emph{no map-space value fusion};
(3) \emph{VISTAv2 (ours)} — short-horizon, action-conditioned imagination projected to an \emph{egocentric value map} and fused at score level.

{\setlength{\textfloatsep}{18pt}
\begin{table}[t]
\centering
\caption{\textbf{Ablation on R2R.} VLFM (semantics only) $\rightarrow$ VISTA (imagination w/o value) $\rightarrow$ VISTAv2 (imagination$\to$value + fusion).}
\label{tab:ablate_vlfm_vista_ours_stacked}
\small
\begin{tabular}{lcccc}
\toprule
\multicolumn{5}{c}{\textbf{Val-Unseen}}\\
\cmidrule(lr){1-5}
Method & TL$\downarrow$ & NE$\downarrow$ & SR$\uparrow$ & SPL$\uparrow$ \\
\midrule
VLFM~\cite{yokoyama2023vlfmvisionlanguagefrontiermaps} & -- & -- & \GThree{52.5} & \GThree{30.4} \\
VISTA~\cite{huang2025vistagenerativevisualimagination} & \GTwo{13.26} & \GOne{2.92} & \GTwo{77.8} & \GTwo{68.3} \\
VISTAv2 (ours)                                          & \GOne{10.73} & \GTwo{3.76} & \GOne{81.4} & \GOne{73.7} \\
\midrule
\multicolumn{5}{c}{\textbf{Test-Unseen}}\\
\cmidrule(lr){1-5}
Method & TL$\downarrow$ & NE$\downarrow$ & SR$\uparrow$ & SPL$\uparrow$ \\
\midrule
VLFM~\cite{yokoyama2023vlfmvisionlanguagefrontiermaps} & --    & --    & \GThree{48.2}   & \GThree{26.4}  \\
VISTA~\cite{huang2025vistagenerativevisualimagination} & \GTwo{14.20} & \GOne{3.77} & \GOne{74.9} & \GTwo{66.7} \\
VISTAv2 (ours)                                          & \GOne{12.44} & \GTwo{4.05} & \GTwo{73.1} & \GOne{69.0} \\
\bottomrule
\end{tabular}
\end{table}

\paragraph{Impact of imagination (VLFM $\to$ VISTA).}
Adding imagination beyond language–vision scoring yields large gains on Val-Unseen: +25.3 SR and +37.9 SPL (77.8/68.3 vs.\ 52.5/30.4), fixing many visually plausible yet unreachable detours. The higher SPL indicates fewer backtracks and more direct paths, consistent with the model pruning dead-ends and avoiding “bait” views (mirrors, glass, long hallways). Qualitatively, short-horizon egocentric rollouts provide explicit \emph{reachability evidence}. For example, whether a corridor bends behind the camera, a doorway is traversable, or a landmark is occluded by walls, frontiers that merely \emph{look} semantically promising no longer dominate the ranking. Notably, this improvement comes \emph{without} additional path supervision: simulated futures act as a geometry-aware prior that complements language matching and stabilizes decisions across scenes.

\begin{figure*}
    \centering
    \includegraphics[width=1\linewidth]{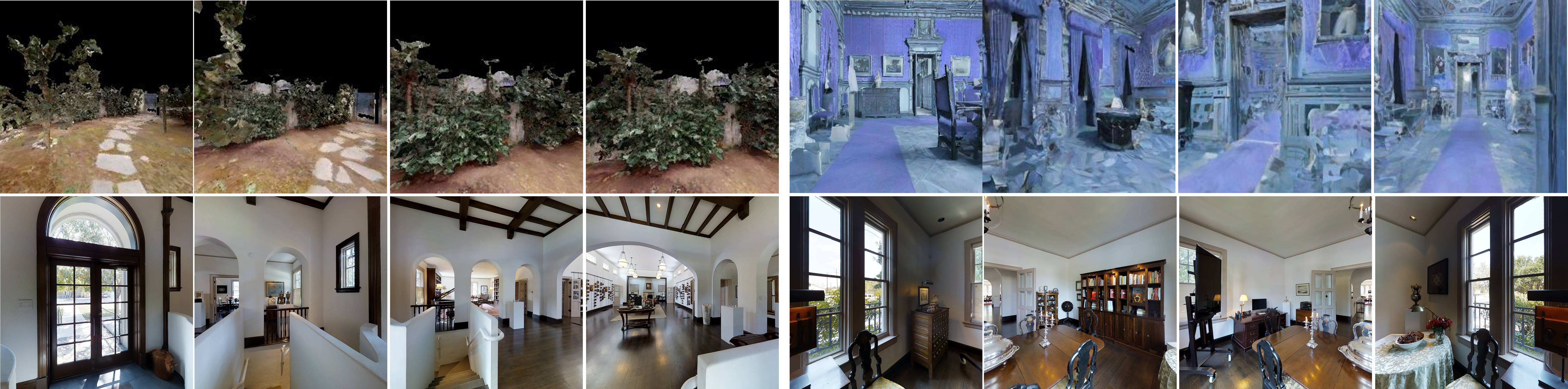}
    \caption{Qualitative visualization of the world-model rollout. For two trajectories in MP3D and HM3D. The rollouts capture room layout and semantics (doorways, arches, windows, tables and bookshelves), which are sufficient for planning even when textures appear stylized.}
    \label{Figure3}
\end{figure*}

\paragraph{Impact of imagination-to-value fusion (VISTA $\to$ VISTAv2).}
Projecting short-horizon rollouts to a \emph{map-space value} and fusing at score level improves both success and efficiency: on Val-Unseen, +1.6 SR, +5.4 SPL, and TL $\downarrow$19.1\%; on Test-Unseen, SPL +2.3 and TL $\downarrow$12.4\%, with a small SR trade-off (\,-1.8) and slightly higher NE (+0.28). Qualitatively, the value prior acts as a \emph{reachability-aware tie-breaker} among candidates with similar base scores, suppressing imagined frontiers and zig-zag oscillations and yielding more direct paths (higher SPL, shorter TL). Because fusion \emph{preserves} the planner, geometric constraints remain intact, while language alignment and traversability cues steer the search toward feasible, instruction-relevant corridors. 
\noindent This ablation isolates the effects: \emph{frontier-only} $\rightarrow$ \emph{imagination} $\rightarrow$ \emph{imagination-to-value}, where the last step yields the largest SPL gains and clear TL reductions.

And we sweep the uncertainty–gating threshold $\theta$ (percentile of rollout variance $\sigma_A$) on R2R Val-Unseen. It exhibits a single-peak trend: too small $\theta$ degenerates to no fusion (VISTA), while too large $\theta$ admits noisy rollouts. We therefore set $\theta{=}0.6$ consistent with our main results (Fig.~\ref{Figure 5}).

\paragraph{Failure case study.}
In scenes with large mirrors or glass partitions, the imagined rollout can overestimate traversability and inflate language alignment behind reflective surfaces. In one episode, the instruction asks the agent to reach a visible landmark in the adjacent room. From the start pose, a full-wall mirror creates a strong false “continuation” of the corridor; monocular depth underestimates the barrier and the world model predicts a plausible forward corridor in a few steps. After projection, the I2V map places a high-value ridge straight ahead, and the fused score ranks this candidate above an alternative that first detours around the corner. As the agent advances, the base mapper eventually detects the blockage, triggering a local oscillation and an early stop just short of the correct doorway—yielding lower SR and slightly higher NE despite a short TL.

{\setlength{\textfloatsep}{8pt}
\begin{figure}[t]
  \centering
  \includegraphics[width=0.86\columnwidth,trim=10 8 10 6,clip]{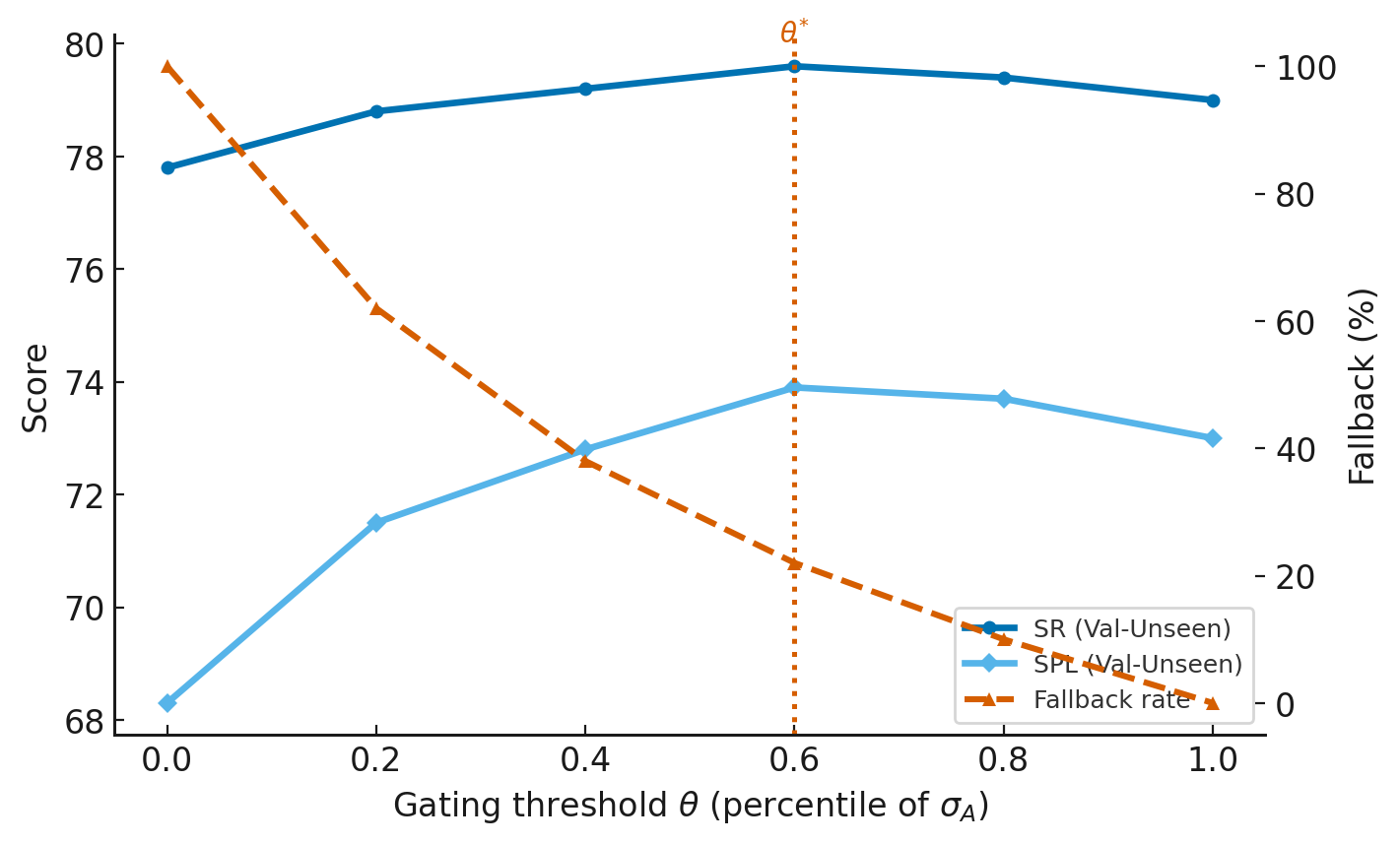}
  \vspace{-2pt}
  \caption{Uncertainty gating sweep on R2R (Val-Unseen). SR/SPL versus the gating threshold $\theta$ (right axis: fallback rate). Performance peaks at a moderate $\theta$ as 0.6.}
  \label{Figure 5}
  \vspace{-4pt}
\end{figure}

%% file: sec/5_conclusion.tex
\section{Conclusion}
\label{sec5.1}
We presented \textbf{VISTAv2}, a language-conditioned, action-aware world model for Vision-and-Language Navigation that rolls out short-horizon egocentric futures and expresses their guidance as an \emph{egocentric value map} in map space. Instead of replacing the planner with a long-horizon objective, VISTAv2 fuses the imagined value \emph{at score level} with a standard frontier based stack, injecting instruction-consistent and reachability-aware cues at test time. On R2R and RoboTHOR, VISTAv2 delivers consistent gains in SR and SPL with shorter paths, and ablations show that imagination helps beyond language priors and projecting imagination into map space value is key to efficiency and robustness. The value maps are interpretable and lightweight to deploy, suggesting a practical route for bringing generative world models into embodied navigation.

%% file: sec/6_limitation.tex
\section{Limitations and Future Work}
Our rollouts are short-horizon and rely on monocular depth and pose estimates; failures occur in low-texture, reflective, or cluttered scenes where uncertainty increases. Future work includes uncertainty-aware planning over longer horizons, richer action spaces, memory for multi-episode goals, and sim-to-real transfer on mobile robots.